\documentclass[sigconf]{acmart}

\usepackage[switch]{lineno}
\usepackage{bm}
\usepackage{multirow}

\AtBeginDocument{%
	\providecommand\BibTeX{{%
			\normalfont B\kern-0.5em{\scshape i\kern-0.25em b}\kern-0.8em\TeX}}}

\copyrightyear{2021}
\acmYear{2021}
\setcopyright{acmcopyright}\acmConference[MM '21]{Proceedings of the 29th ACM
International Conference on Multimedia}{October 20--24, 2021}{Virtual Event, China}
\acmBooktitle{Proceedings of the 29th ACM International Conference on Multimedia
(MM '21), October 20--24, 2021, Virtual Event, China}
\acmPrice{15.00}
\acmDOI{10.1145/3474085.3475302}
\acmISBN{978-1-4503-8651-7/21/10}

\begin{document}
\fancyhead{}

\title{Cut-Thumbnail: A Novel Data Augmentation \\ for Convolutional Neural Network}


\author{Tianshu Xie}
\authornote{Both authors contributed equally to this research.}
\email{tianshuxie@std.uestc.edu.cn}
\affiliation{%
  \institution{University of Electronic Science and Technology of China}
  \city{Chengdu}
  \country{China}
}

\author{Xuan Cheng}
\authornotemark[1]
\email{cs_xuancheng@std.uestc.edu.cn}
\affiliation{%
  \institution{University of Electronic Science and Technology of China}
   \city{Chengdu}
  \country{China}
}

\author{Xiaomin Wang}
\authornote{Xiaomin Wang is the corresponding author.}
\email{xmwang@uestc.edu.cn}
\affiliation{%
  \institution{University of Electronic Science and Technology of China}
   \city{Chengdu}
  \country{China}
}

\author{Minghui Liu}
\email{minghuiliuuestc@163.com}
\affiliation{%
  \institution{University of Electronic Science and Technology of China}
   \city{Chengdu}
  \country{China}
}

\author{Jiali Deng}
\email{julia_d@163.com}
\affiliation{%
  \institution{University of Electronic Science and Technology of China}
   \city{Chengdu}
  \country{China}
}

\author{Tao Zhou}
\email{zhou_tao@uestc.edu.cn}
\affiliation{%
  \institution{University of Electronic Science and Technology of China}
   \city{Chengdu}
  \country{China}
}

\author{Ming Liu}
\email{csmliu@uestc.edu.cn}
\affiliation{%
  \institution{University of Electronic Science and Technology of China}
   \city{Chengdu}
  \country{China}
}

\renewcommand{\shortauthors}{Trovato and Tobin, et al.}

\begin{abstract}
In this paper, we propose a novel data augmentation strategy named Cut-Thumbnail, that aims to improve the shape bias of the network. We reduce an image to a certain size and replace the random region of the original image with the reduced image. The generated image not only retains most of the original image information but also has global information in the reduced image. We call the reduced image as thumbnail. Furthermore, we find that the idea of thumbnail can be perfectly integrated with Mixed Sample Data Augmentation, so we put one image's thumbnail on another image while the ground truth labels are also mixed, making great achievements on various computer vision tasks. Extensive experiments show that Cut-Thumbnail works better than state-of-the-art augmentation strategies across classification, fine-grained image classification, and object detection. On ImageNet classification, ResNet-50 architecture with our method achieves 79.21\% accuracy, which is more than 2.8\% improvement on the baseline.
\end{abstract}

\ccsdesc[500]{Computing methodologies}
\ccsdesc[500]{Computing methodologies~Neural networks}
\ccsdesc[500]{Computing methodologies~Supervised learning by classification}

\keywords{classification, convolutional neural network, thumbnail, data augmentation, regularization}

\maketitle

\section{Introduction}
In recent years, the deep convolutional neural network (CNN) has made remarkable achievements in computer vision, including image classification~\cite{krizhevsky2012imagenet,russakovsky2015imagenet,szegedy2015going,he2016deep}, object detection~\cite{girshick2015fast,ren2015faster,he2017mask}, and semantic segmentation~\cite{long2015fully,chen2017deeplab}. However, its huge structure and massive parameters pose a challenge to the training of the network. Many data augmentation and regularization approaches have been proposed to solve this problem.

As an important technology to generate more useful data from existing ones, data augmentation can significantly enhance network performance. The most commonly used data augmentation methods are spatial transformations, including random scale, crop, flip and random rotation~\cite{krizhevsky2012imagenet}. Cutout~\cite{devries2017improved} and Random Erasing~\cite{zhong2020random} randomly set black blocks or place noises in one or more areas. Color distortion~\cite{szegedy2015going} changes the brightness of training images. Mixup~\cite{zhang2017mixup} and CutMix~\cite{yun2019cutmix} combines the two images with different strategies, and two images' labels are also mixed. In a word, the most existing data augmentation techniques improve the generalization ability and robustness of the network by changing spatial or color information, adding noise, or mixing information from different images.

\begin{figure}[t]
\centering
\includegraphics[width=0.95\columnwidth]{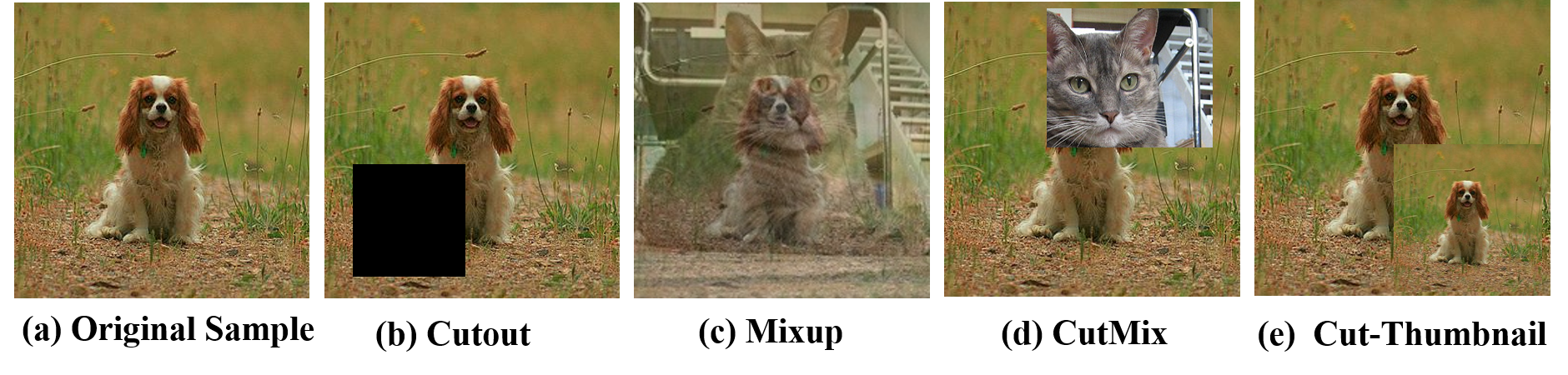} 
\caption{Comparison between existing data augmentation methods with Cut-Thumbnail.}
\label{fig1}
\end{figure}

In this paper, we introduce the idea of thumbnail into data augmentation and propose a novel augmentation strategy named Cut-Thumbnail. Figure~\ref{fig1} illustrates the comparison between existing data augmentation methods and Cut-Thumbnail. We reduce an image to a small size thumbnail and replace the original image's random area with it. Though being reduced, the thumbnail still contains most semantic information of the original image. By doing this, we not only make the network learn the features of images with different sizes, but also strengthen the network's capture of shape information. Furthermore, Cut-Thumbnail can be perfectly integrated with Mixed Sample Data Augmentation (MSDA), which refers to the combination of data samples according to a certain strategy. It is because when using a thumbnail to replace another image's random area, the thumbnail can completely contain the global information of its original image without taking up much semantics of another image. Therefore, we use a thumbnail to replace another image's random region, and mix their labels with certain weights. Besides, we find introducing two or more thumbnails into another image can also improve the network's effect on specific datasets. We specify a series of strategies around Cut-Thumbnail that will be presented in Section 3.

To demonstrate Cut-Thumbnail's effectiveness, we conduct extensive experiments on various CNN architectures, datasets, and tasks. On ImageNet~\cite{russakovsky2015imagenet}, Cut-Thumbnail can improve the accuracy of ResNet-50~\cite{he2016deep} from 76.32\% to 79.21\%, more effective than state-of-the-art method CutMix, which accomplishes 78.40\%. On CIFAR100~\cite{krizhevsky2009learning}, applying Cut-Thumbnail to ResNet-56  and WideResNet-28-10~\cite{zagoruyko2016wide} has improved the classification accuracy by +3.07\% and +2.45\%, respectively. Furthermore, On the CUB-200-2011~\cite{wah2011caltech} dataset for the fine-grained classification task, Cut-Thumbnail increases the accuracy of ResNet-50 from 85.31\% to 87.76\%. On the Pascal VOC~\cite{everingham2010pascal} dataset for the object detection task, our method increases the mAP of RetinaNet~\cite{lin2017focal} from 70.14\% to 72.16\%.

To sum up, this paper makes the following contributions:
\begin{itemize}
\item We propose Cut-Thumbnail, a simple but effective data augmentation strategy \textit{first} introducing the idea of thumbnail to data augmentation, which aims to make the network better learn the shape information.
\item We combine Cut-Thumbnail with MSDA, the generated images of Thumbail are natural and contain most semantics of the reduced image.
\item We conduct extensive experiments on classification, fine-grained classification and object detection. Comparing with state-of-the-art data augmentation methods, the experimental results demonstrate that our method achieves the best performance.
\end{itemize}

\section{Motivation}
\noindent \textbf{Shape v.s Texture:}
Recent studies have shown that CNN is texture-biased, $i.e.$ CNN relys more on local texture rather than global shape in decision-making~\cite{geirhos2018imagenet,gatys2017texture,brendel2019approximating,ballester2016performance,shi2020informative}. CNN can classify texture images well, but it is not sensitive to the shape of objects. For example, CNN tends to classify an image with a cat shape filled with an elephant skin texture as an elephant instead of a cat~\cite{geirhos2018imagenet}. ~\cite{geirhos2018imagenet,shi2020informative} denote that improving the shape bias of CNN can improve the accuracy and robustness. However, the shape information contained in image is scarce and vague, making the network difficult to capture effective shape information.

\begin{figure}[h]
\centering
\includegraphics[width=0.95\columnwidth]{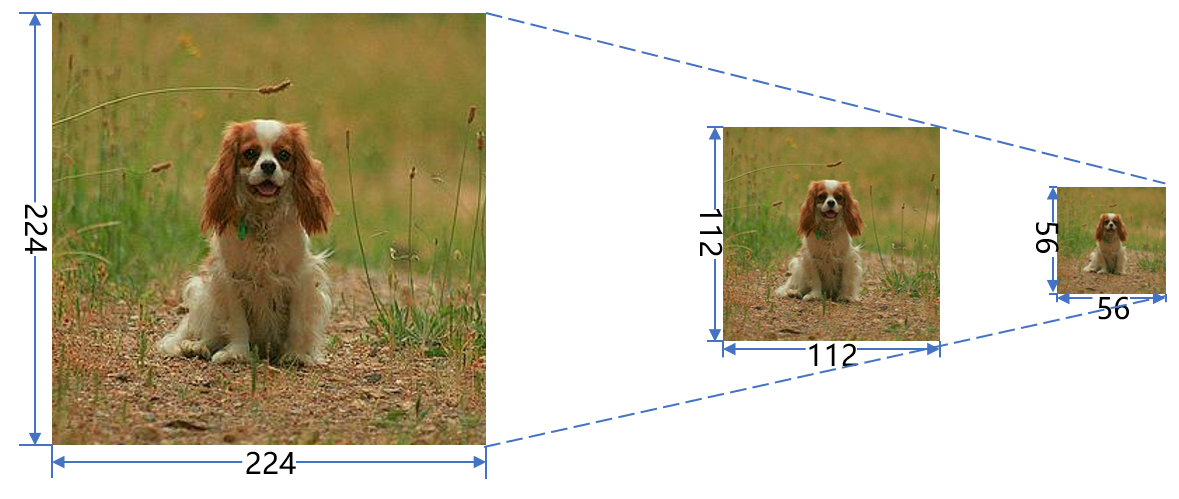} 
\caption{This image shows an example of reduced images that we call thumbnails. After reducing the image to a certain size 112$\times$112 or 56$\times$56, we can still recognize a dog in the image even though lots of local details are lost.}.
\label{fig2}
\end{figure}

\begin{figure*}[t]
\centering
\includegraphics[width=1.98\columnwidth]{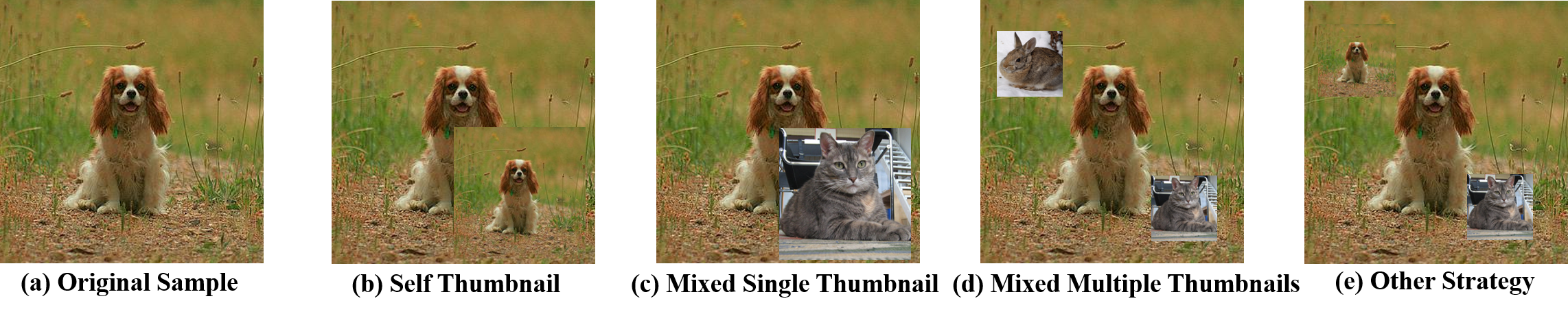}
\caption{Illustration of Cut-Thumbnail. We put a single thumbnail or multiple thumbnails on the thumbnail's original image or another image, and thus get different strategies.}
\label{fig3}
\end{figure*}

Different from~\cite{geirhos2018imagenet} which consumes lots of computational overhead to generate images with little texture information, we push forward a novel data augmentation strategy to force the network to perceive more shape information by utilizing the property of thumbnail. As shown in Figure~\ref{fig2}, we can still recognize that there is a dog in the image after reducing the image to a certain size. We call the reduced image as a thumbnail. Although lots of the texture details are lost, we can still identify the target in the thumbnail. This is because human usually relies on the global information of the image~\cite{landau1988importance,diesendruck2003specific}, such as shape, for image classification, which is still reserved by the thumbnail. We consider using the shape information that the thumbnail contained is a simple but effective way to improve the shape bias of network. When we put the thumbnail on its original image, the network can learn both the shape information and the texture information simultaneously. To verify whether the network trained with Cut-Thumbnail can improve the shape bias of CNN, we conduct an experiment that can be seen in SubSection 3.4. The result shows that our network performs better on the dataset with the grayscale images that contain more shape information and less texture information than other methods.

\noindent \textbf{Cut-Tumbnail v.s CutMix:}
While both replacing the image region, CutMix~\cite{yun2019cutmix} tends to use a random patch from another image, which is quite possible to be uninformative. Cut-Tumbnail overcomes this problem by using the thumbnail, which retains most semantics of another image. Note that CutMix uses the patch of the image with no complete shape information, and just strengths the texture learning of CNN. While Cut-Thumbnail uses thumbnail that contains complete shape information and improves the shape bias of CNN. As can be seen in later experiments, this strategy significantly improves network performance and gets better results in classification, fine-grained classification, and object detection than CutMix.

\section{Our Approach}
Cut-Thumbnail is an effective albeit simple data augmentation technique for CNN. For a given training sample $(x,y)$ which $x \in \mathbb R^{W\times H\times C} $ denotes the training image and $y$ denotes the training label, we get a thumbnail image $T(x)$ by simply taking one pixel out of a certain number of pixels of the image $x$. We put a single thumbnail or multiple thumbnails on the thumbnail's original image or another image, and thus get different strategies. Next, we will introduce them in turn.

\subsection{Self Thumbnail (ST)}

In our first strategy, we use the thumbnail $T(x)\in \mathbb R^{w\times h\times C}$ to replace a random region of the original image $x$  and do not change the label, which is called Self Thumbnail as shown in Figure~\ref{fig3}(b). For a given training sample $(x_1,y_1)$, We define this operation as

\begin{equation}
\begin{aligned}
\tilde{x} &= \bm{{\rm M}} \odot x_1 + \Phi (T(x_1)) \\
\tilde{y} &= y_1
\end{aligned}
\end{equation}
where $(\tilde{x},\tilde{y})$ denote the generated sample, $\bm{{\rm M}} \in \{0,1\}^{W\times H} $ is the binary mask indicating where to drop out and fill in from origin image and thumbnail, and $\odot$ is element-wise multiplication. To sample the binary mask $\bm{{\rm M }}$, we first sample the bounding box coordinates $\bm{{\rm B }} = (r_x,r_y,r_w,r_h)$ indicating the cropping regions on $x_1$. The region $\bm{{\rm B }}$ in $x_1$ is removed and filled in with the thumbnail $T(x_1)$.  The box coordinates are uniformly sampled according to

\begin{equation}
\begin{aligned}
&r_x \sim {\rm Unif}(0,W), r_w = w \\
&r_y \sim {\rm Unif}(0,H), r_h = h
\end{aligned}
\end{equation}
where $w,h$ denote the width and height of the thumbnail $T(x_1)$, which are usually set to half the width and height of the original image. With the cropping region, the binary mask $\bm{{\rm M }}$ is decided by filling with 0 within the bounding box $\bm{{\rm B }}$, otherwise 1. $\Phi(\cdot)$  denotes the padding operation that generates an image with the same size as $x_1$. $\Phi$ first generates a binary mask ${\bm{{\rm  \tilde M }}} = \bm{1} -  \bm{{\rm M}} $, $\bm{1}$ is a binary mask filled with ones. The bounding box coordinates $\bm{{\rm B }}$ still exists in $\bm{{\rm \tilde{ M }}}$, so we put the thumbnail $T(x_1)$ in $\bm{{\rm B }}$ and the generated image is $\Phi(T(x_1))$. This strategy enables the network to learn the same image at different scales. In addition to the information obtained from the original image, the thumbnail can provide the global information for the training, which plays a guiding role in the network learning.

\subsection{Mixed Single Thumbnail (MST)}
The idea of thumbnail is very suitable for Mixed Sample Data Augmentation (MSDA), which involves combining data samples according to a certain policy to create an augmented data set. In our second strategy, one image's random region is replaced by another's thumbnail where their labels are multiplied by different weights and added, so that most of the generated images contain the information of two images as shown in Figure~\ref{fig3}(c). We call this strategy as Mixed Single Thumbnail. For a pair of given training sample$(x_1,y_1)$ and $(x_2,y_2)$, we define this combining operation as

\begin{equation}
\begin{aligned}
\tilde{x} &= \bm{{\rm M}} \odot x_1 + \Phi (T(x_2)) \\
\tilde{y} &= (1-\lambda) y_1 + \lambda y_2
\end{aligned}
\end{equation}
where $\bm{{\rm M }}$, $\Phi$, and the size of thumbnail $T(x_2)$ is set in the same way as Self Thumbnail. This strategy combines the idea of thumbnail with MSDA, so that the network can learn the original information of one image and the complete information of another image simultaneously.

\subsection{Mixed Multiple Thumbnails (MMT)}
One of thumbnail's advantages is that it can introduce one image's complete semantics by occupying a small area of another image. Besides, unlike the simple overlay in Mixup, Cut-Thumbnail does not make the original image appear unnatural. Therefore, we can further expand Cut-Thumbnail's superiority by putting two or more thumbnails on another image as shown in Figure~\ref{fig3}(d). Take adding $n$ thumbnails to another image as an example, for given training samples $(x_1,y_1)$ , $(x_2,y_2)$ , ... , $(x_n,y_n)$, we define this combining operation as

\begin{equation}
\begin{aligned}
\tilde{x} &= \bm{{\rm M}} \odot x_1 +   \sum_{i=2}^{n}\Phi_i (T(x_i))  \\
\tilde{y} &= \lambda_1 y_1 + \sum_{i=2}^{n}\lambda_i y_i
\end{aligned}
\end{equation}
where $\bm{{\rm M }}$ contains $n-1$ bounding box coordinates $\bm{{\rm B_2}}$, $\bm{{\rm B_3}}$, ... , $\bm{{\rm B_{n}}}$ corresponding to $\Phi_2$, $\Phi_3$, ... , $\Phi_n$. Unlike CutMix, the weight of image labels $\lambda_i$ is not determined by the area of the thumbnails. It is because CutMix only adds the random parts of two images, but the thumbnail has most of the original image's semantics, it should have a higher weight, which can be seen in Section 4.5. Mixed Multiple Thumbnails can make each training image contain more images' global information, improving the network training efficiency. We find that Mixed Multiple Thumbnails significantly improves the network performance on datasets with the larger image size and less data volume like CUB-200-2011~\cite{wah2011caltech}.

\noindent During training, we randomly choose 80\% of batches using Cut-Thumbnail rather than every batch. We call the rate of batches applying Cut-Thumbnail as $participation\_rate$. Note that although our method has achieved a high level of improvement, we have not deliberately sought for the optimal combination of these strategies due to the limitation of time and computing resources. In other words, the potential of Cut-Thumbnail can be explored in future work.

\subsection{Why Does Cut-Thumbnail Help?}
\noindent \textbf{Graysacle image recognition:} The shape information contained in thumbnail, together with the texture bias of CNN, further motivates our proposed method that aims to improve the shape bias of CNN. To verify that Cut-Thumbnail can indeed make the network learn more shape information, we use grayscale images as the test set to compare the performance of the networks trained by different methods. Due to the lack of color information, graysacle images have less texture details, but its shape information is not affected. Therefore, the recognition of grayscale image requires the network to rely more shape information compared with color images.

\begin{table}[h]
  \begin{center}
  \begin{tabular}{lc}
  \toprule[1.2pt]
  \midrule
       Model&Greyscale Image(\%)  \\
        \hline
       ResNet-50(Baseline) &64.70 \\
       ResNet-50+CutMix  &67.61(+2.91) \\

       ResNet-50+ST  &66.95(+2.25) \\

	   ResNet-50+MST &{\bfseries 68.63(+3.93)} \\
  \midrule
  \bottomrule[1.2pt]
  \end{tabular}
  \end{center}
  \caption{Comparison of CutMix and Cut-Thumbnail on greyscale image with ResNet-50. ST denotes Self Thumbnail and MST denotes Mixed Single Thumbnail.}
  \label{tab0}
\end{table}

We transform all the images in ImageNet's verification set into grayscale images as a new verification set. The tested networks are trained on regular Imagenet training set with different methods, and the training details can be seen in SubSection 4.1. As demonstrated in Table~\ref{tab0}, The network trained with Mixed Single Thumbnail(MST) has achieved the highest results(+3.93\%) on  grayscale image. The performance on grayscale image of our method shows that the shape information contained in the thumbnail is helpful for improving the shape bias of CNN. In the absence of texture information, the grayscale image recognition of network trained by MST is better than CutMix, which shows that our method can enhance the learning of image shape.

\begin{figure}[h]
\centering
\includegraphics[width=0.95\columnwidth]{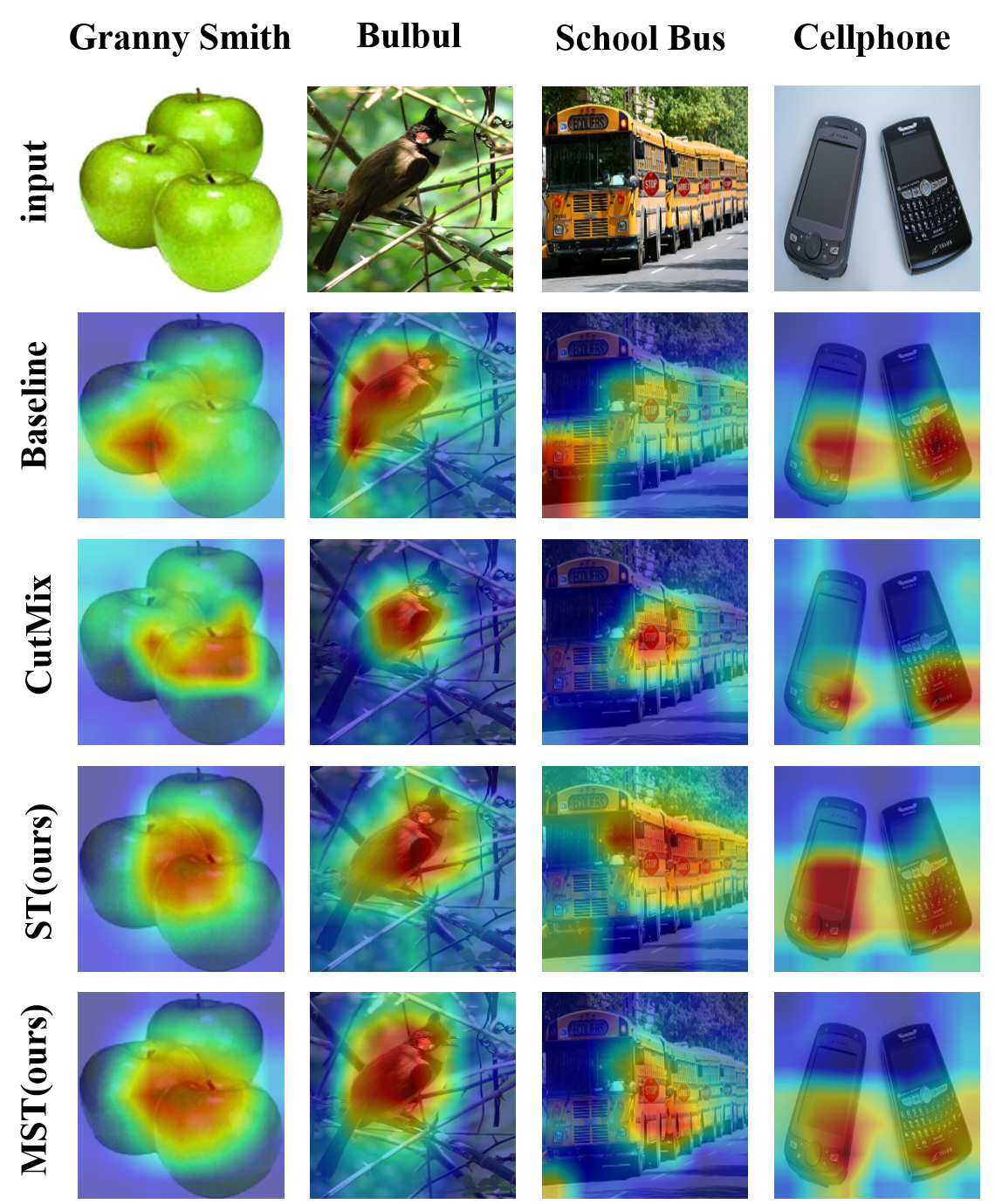} 
\caption{Class activation mapping (CAM)~\cite{zhou2016learning} for ResNet-50 model on ImageNet, with baseline augmentation, CutMix, ST or MST.}
\label{fig8}
\end{figure}

\noindent \textbf{Network visualization:} To analyze what the model trained with Cut-Thumbnail learns, we compute class activation mapping (CAM) for ResNet-50 model trained with ST and MST on ImageNet. We also show the CAM for models trained with baseline augmentation and CutMix for comparison in Figure~\ref{fig8}. The models trained with ST and MST both tend to focus on large important regions, while the network trained by CutMix tends to focus on local regions. This proves that the network trained with Cut-Thumbnail is biased to the global information of the image for object recognition, which is different from CutMix that makes the network pay more attention to the local regions.

\section{Experiment}
In this section, we investigate the effectiveness of Cut-Thumbnail for several major computer vision tasks. We first conduct extensive experiments on image classification and fine-grained image classification. Next, we study the effect of Cut-Thumbnail on object detection. All experiments are performed with Pytorch~\cite{paszke2017automatic} on Tesla M40 GPUs.
\subsection{ImageNet Classification}

ImageNet-1K~\cite{russakovsky2015imagenet} contains 1.2M training images and 50K validation images labeled with 1K categories. We use the standard augmentation setting for ImageNet dataset such as resizing, cropping, and flipping. For fair comparison, the model is trained from scratch for 300 epochs with batch size 256 and the learning rate is decayed by the factor of 0.1 at epochs 75, 150, 225, as done in CutMix~\cite{yun2019cutmix}. We evaluate classification accuracy on the validation set and the highest validation accuracy is reported over the full training course following the common practice. For Self Thumbnail (ST) and Mixed Single Thumbnail (MST), we set the $thumbnail\_size$ to 112$\times$112 which is half of the image width and height, and $\lambda$ in MST is set to 0.25. We explore the performance of different data augmentation methods on ResNet-18 and ResNet-50~\cite{he2016deep}. The results are illustrated on Table~\ref{tab1}.

\noindent \textbf{Performance on ResNet-18:} With ST, we improve the accuracy of ResNet-18 from 70.10\% to 71.92\% (+1.82\%), which surpasses CutMix significantly. The improvement of CutMix on ResNet-18 is not obvious, we speculate that it is because the images generated by CutMix are relatively complex for ResNet-18 with weak learning ability. ST uses the image's own thumbnail to replace its random region without adding the extra image information, which may be more beneficial to the training of small networks like ResNet-18.

\begin{table}[t]
  \begin{center}
  \begin{tabular}{llc}
  \toprule[1.2pt]
  \midrule
  Model&Method&Accuracy(\%)\\\hline
·\multirow{4}{*}{ResNet-18} & baseline&69.90$\pm$0.09 \\
                    & +CutMix&70.30$\pm$0.03  \\
                    &+ST (ours)&{\bfseries 71.92$\pm$0.04} \\
                    &+MST (ours)&71.34$\pm$0.07\\
  \midrule
  \multirow{8}{*}{ResNet-50} & baseline&76.32$\pm$0.02\\
		&+Cutout&77.07$\pm$0.04\\
        &+Mixup&77.42$\pm$0.06\\
        &+AutoAugment*&77.63\\
        &+DropBlock*&78.13$\pm$0.05\\
        &+CutMix&78.40$\pm$0.04\\
		&+ST (ours)&77.74$\pm$0.05\\
		&+MST (ours)&{\bfseries 79.21$\pm$0.04}\\
  \midrule
  \bottomrule[1.2pt]
  \end{tabular}
  \end{center}
  \caption{Summary of validation accuracy of the ImageNet classification results based on ResNet-18 and ResNet-50. ST denotes Self Thumbnail and MST denotes Mixed Single Thumbnail. We report average over 3 runs. `*' means results reported in the original paper.}
  \label{tab1}
\end{table}

\noindent \textbf{Performance on ResNet-50:} ResNet-50 is a widely used CNN architecture for image recognition. We can observe that MST achieves the best result, 79.21\% top-1 accuracy, among the considered augmentation strategies.
Cutout~\cite{devries2017improved} randomly masks square sections of the image. We set the mask size for Cutout to $112 \times 112$ and the location for dropping out is uniformly sampled. Inspired by Cutout, DropBlock~\cite{ghiasi2018dropblock} randomly drops some contiguous regions of a feature map. MST outperforms Cutout and DropBlock by +2.14\% and +1.08\%, respectively.

Mixup and CutMix are the successful variants of MSDA, which achieve excellent results in classification tasks.  We set $\alpha = 1$ in both Mixup~\cite{zhang2017mixup} and CutMix. MST outperforms Mixup and CutMix,  by +1.79\% and +0.81\%, respectively. It shows that the images generated with thumbnail are more conducive to network learning.

Note that ST also achieves a great performance on ResNet-50. ST utilizes only one image's information, but it performs better than Mixup using multiple images' information. Besides, AutoAugment~\cite{cubuk2018autoaugment} uses reinforcement learning to find a combination of existing augmentation policies. ST, by simply putting the image's own thumbnail on itself, even exceeds the performance of AutoAugment, which demonstrates the effectiveness and generality of our method.

\begin{figure}[h]
\centering
\includegraphics[width=0.98\columnwidth]{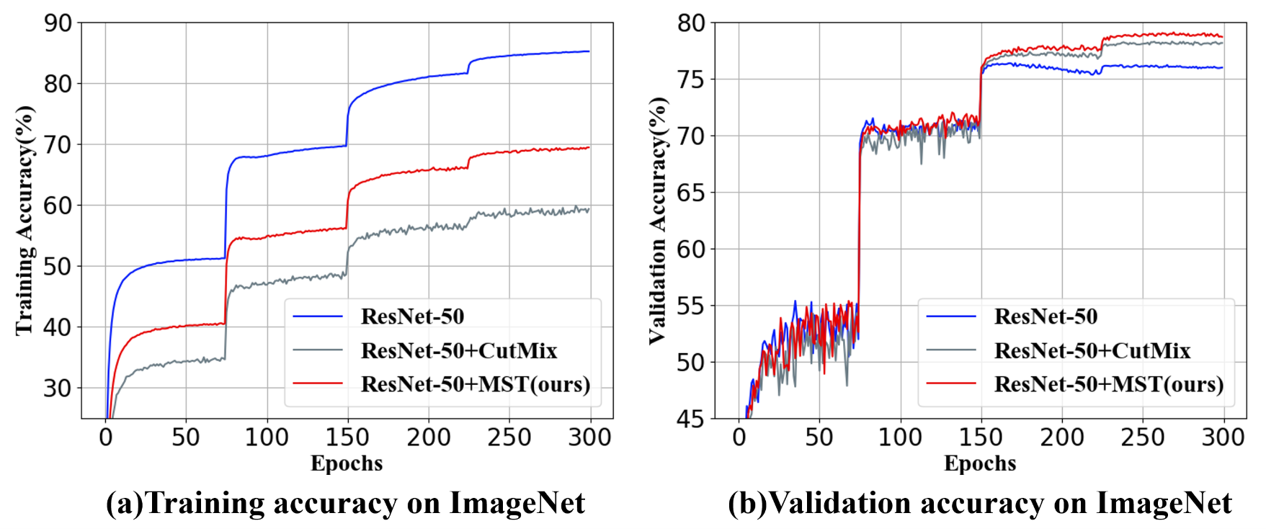} 
\caption{ Training and validation accuracy comparison among baseline, CutMix and MST on ImageNet with ResNet-50.}
\label{fig4}
\end{figure}

Training and validation accuracy comparison among baseline, CutMix and MST on ImageNet with ResNet-50 can be seen in Figure~\ref{fig4}. Due to the mixing of labels and images, the accuracy of MSDA methods such as Mixup and CutMix is far lower than the baseline, and MST is no exception. But the validation accuracy of CutMix and MST is much higher than baseline and the accuracy of MST is higher than that of CutMix, which shows that our method can significantly improve the generalization of the network.

\subsection{Tiny ImageNet Classification}
Tiny ImageNet dataset is a subset of the ImageNet dataset with 200 classes. Each class has 500 training images, 50 validation images, and 50 test images. All images are with $64\times64$ resolution. We test the performance of ResNet-110 on this dataset. The learning rate is initially set to 0.1 and decayed by the factor of 0.1 at epochs 150 and 225. For ST and MST, we set the size of the thumbnail to 32$\times$32, and $\lambda$ in MST is set to 0.25. The hole size of Cutout is set to 32$\times$32. For Mixup and CutMix, the hyper-parameter $\alpha$ is set to 1.0. The results are summarized on Table~\ref{tab2}. MST achieves the best performance 66.45\% on Tiny ImageNet. This proves that our method is also generalized for datasets with different data sizes.

\begin{table}[h]
  \begin{center}
  \begin{tabular}{p{4.5cm}p{1.8cm}}
  \toprule[1.2pt]
  \midrule
        Model&Accuracy(\%)\\\hline
        ResNet-110 (baseline)&62.42$\pm$0.02\\
		ResNet-110+Cutout&64.71$\pm$0.09\\
        ResNet-110+Mixup&65.34$\pm$0.14\\
        ResNet-110+CutMix&66.13$\pm$0.02\\
  \midrule
		ResNet-110+ST (ours)&64.85$\pm$0.04\\
		ResNet-110+MST (ours)&{\bfseries 66.45$\pm$0.03}\\
  \midrule
  \bottomrule[1.2pt]
  \end{tabular}
  \end{center}
  \caption{Comparison of state-of-the-art data augmentation methods on Tiny ImageNet with ResNet-110.}
  \label{tab2}
\end{table}

\begin{figure*}[t]
\centering
\includegraphics[scale=0.47]{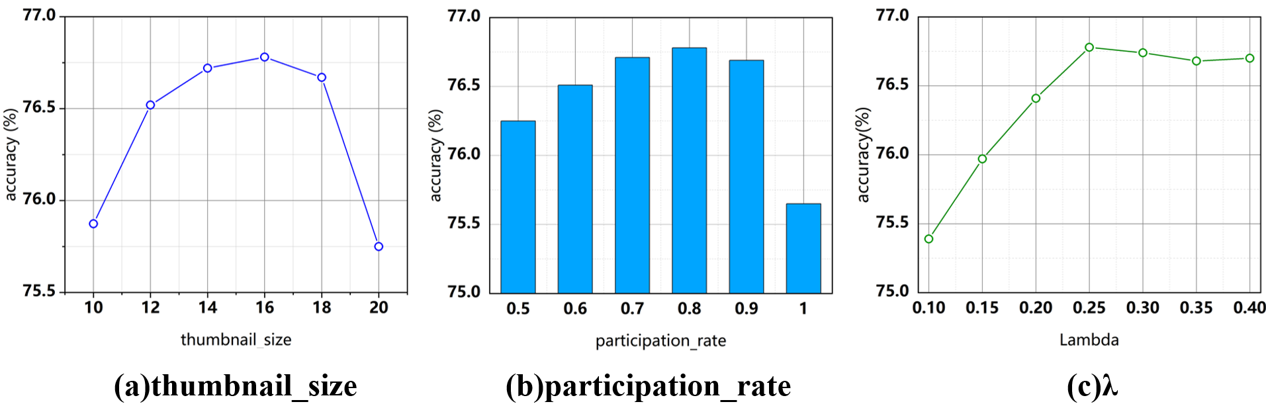}
\caption{CIFAR100 validation accuracy against \bm{$thumbnail\_size$}, \bm{$particitpation\_rate$}, and \bm{$\lambda$} with ResNet-56. We set \bm{$thumbnail\_size=16$}, \bm{$particitpation\_rate=0.8$}, and \bm{$\lambda=0.25$} as the default parameter settings. }.
\label{fig5}
\end{figure*}

\subsection{CIFAR Classification}
The CIFAR10~\cite{krizhevsky2009learning} dataset collects 60,000 32$\times$32 color images of 10 classes, each with 6000 images including 5,000 training images and 1,000 testing images. The CIFAR100~\cite{krizhevsky2009learning} dataset has the same number of images but 100 classes. To test the universality of our method under different network structures, ResNet-56 and WideResNet-28-10 are selected as baseline. For WideResNet-28-10, the learning rate is decayed by the factor of 0.1 at epochs 60, 120, 160; for ResNet-56, the learning rate is decayed by the factor of 0.1 at epochs 150, 225. For ST and MST, we set the $thumbnail\_size$ to 16$\times$16, and $\lambda$ in MST is set to 0.25. The hole size of Cutout is set to 16$\times$16. For Mixup and CutMix, the hyper-parameter $\alpha$ is set to 1.0.

\begin{table}[h]
  \begin{center}
  \begin{tabular}{lp{2cm}cp{3cm}cp{2cm}}
  \toprule[1.2pt]
  \midrule
  Model&Method&Accuracy(\%)\\\hline
 ·\multirow{6}{*}{ResNet-56}& baseline&73.71$\pm$0.12 \\
        &+Cutout&74.64$\pm$0.15\\
		&+Mixup&75.97$\pm$0.26\\
		&+CutMix&76.57$\pm$0.13\\
        &+ST (ours)&75.58$\pm$0.11\\
        &+MST (ours)&{\bfseries 76.78$\pm$0.08}\\
  \midrule
  \multirow{6}{*}{WideResNet-28-10} & baseline&81.06$\pm$0.03\\
	    &+Cutout&81.86$\pm$0.08\\
		&+Mixup&82.57$\pm$0.12\\
		&+CutMix&83.13$\pm$0.06\\
        &+ST (ours)&81.41$\pm$0.04\\
        &+MST (ours)&{\bfseries 83.35$\pm$0.05}\\
  \midrule
  \bottomrule[1.2pt]
  \end{tabular}
  \end{center}
  \caption{Comparison of Top-1 accuracy of ResNet-56 and WideResNet-28-10 on the CIFAR100 validation set. MST obtains the best performance on both networks. }
  \label{tab3}
\end{table}

As shown in Table~\ref{tab3}, on CIFAR100 dataset, MST provides better results over ResNet-56 and WideResNet-28-10 compared to Cutout, Mixup and CutMix. For ResNet-56, MST achieves a significant 3.07\% improvement over the base model. ST outperforms Cutout on ResNet-56, showing that pasting the image's own thumbnail is better than black block. Generalization is an essential property of data augmentation methods, experiments show that our method is suitable for networks with different structures.

\begin{table}[h]
  \begin{center}
  \begin{tabular}{p{4.5cm}p{1.8cm}}
  \toprule[1.2pt]
  \midrule
  Model&Accuracy(\%)\\\hline
  ResNet-56 (baseline)& 94.00$\pm$0.14\\
  ResNet-56+Cutout& 94.80$\pm$0.18\\
  ResNet-56+Mixup&  95.01$\pm$0.16\\
  ResNet-56+CutMix& {\bfseries 95.33$\pm$0.11}\\
  \midrule
  ResNet-56+ST (ours)&95.03$\pm$0.09\\
  ResNet-56+MST (ours)&95.24$\pm$0.12\\
  \midrule
  \bottomrule[1.2pt]
  \end{tabular}
  \end{center}
  \caption{Impact of Cut-Thumbnail on CIFAR10 for ResNet-56.}
  \label{tab4}
\end{table}

As shown in Table~\ref{tab4}, on CIFAR10 dataset, MST improves the performance by +1.24\% on ResNet-56, but slightly lower than 1.33\% of CutMix. We consider it may be that images in CIFAR10 are relatively simple with low pixels, the information provided by thumbnails is limited. But for CIFAR10 and CIFAR100 with images of very small sizes, our method can also achieve significant performance improvement, which demonstrates Cut-Thumbnail applies to various types of datasets.

\subsection{Ablation Studies}
We conduct ablation studies on CIFAR100 dataset using the same experimental settings of ResNet-56 in Subsection 4.3.

\noindent \textbf{Analysis on \bm{$thumbnail\_size$}:}  We evaluate Cut-Thumbnail with $thumbnail\_size$ $\in$ \{10,12,14,16,18,20\}. As shown in Figure~\ref{fig5}(a), with the increasing of $thumbnail\_size$, the accuracy first rises and then decreases after reaching the highest when the $thumbnail\_size$ is 16, which is half of the image width or height. It denotes that the small thumbnail does not have enough semantics to guide network training, while the large one affects the semantics of the original image. Therefore, we generally select the thumbnail with the half width and height of the original image.

\noindent \textbf{Effect of the \bm{$participation\_rate$}:} Specially, we call the ratio of batches using Cut-Thumbnail to all batches as the $participation\_rate$. As shown in Figure~\ref{fig5}(b), when the participation\_rate  between 0.7 and 0.9, the difference in network performance is not obvious, but they are significantly better than the performance when the $participation\_rate$ is 1. This shows that our method is not sensitive to $participation\_rate$ as long as the $participation\_rate$ is higher than 0.7 and not to be 1. We consider the reason may be that the network needs to supplement a small number of normal images for comparative learning with thumbnail.

\noindent \textbf{Exploration to \bm{$\lambda$}:} We test the effect of $\lambda$ on the training, which is the weight multiplied by the image label. The results are given in Figure~\ref{fig5}(c), when the $\lambda$ is set to 0.25, the model performance is the best. Besides, the difference in network performance is  not obvious either when the $\lambda$ is between 0.25 and 0.35. We consider it because the thumbnail contains most semantics of the original image, so multiplying the thumbnail's label with a higher weight is beneficial to network training.

\subsection{Fine-grained Image Classification}

The fine-grained image classification aims to recognize similar subcategories of objects under the same basic-level category. The difference of fine-grained recognition compared with general category recognition is that fine-grained subcategories often share the same parts and usually can only be distinguished by the subtle differences in texture and color properties of these parts. CUB-200-2011~\cite{wah2011caltech} is a widely-used fine-grained dataset which consists of images in 200 bird species. There are about 30 images for training for each class.

To verify the generalization of different types of computer vision tasks, we use ResNet-50 to test the performance of our method on CUB-200-2011. The training starts from the model pretrained on ImageNet. The mini-batch size is set to 16 and the number of training epoch is set to 95. The learning rate is initially set to 0.001 and decayed by the factor of 0.1 at epochs 30, 60 and 90. During network training, the input images are randomly cropped to 448$\times$448 pixels after being resized to 600$\times$600 pixels and randomly flipped. As shown in Table \ref{tab5}, with Mixed Double Thumbnail (MDT) which denotes two images' thumbnails are put on another image, we improve the accuracy of ResNet-50 from 85.31\% to 86.72\%(+1.41\%), which surpasses previous data augmentation methods significantly. It also shows that the shape information has a high gain effect even on fine-grained image classification which requires more subtle differences.

\begin{table}[h]
	\begin{center}
		\begin{tabular}{p{4.5cm}p{1.8cm}}
			\toprule[1.2pt]
			\midrule
			Model&Accuracy(\%)\\
			\midrule
			ResNet-50 (baseline)&85.31$\pm$0.21\\
			ResNet-50+Cutout&85.68$\pm$0.13\\
			ResNet-50+Mixup&85.91$\pm$0.31\\
			ResNet-50+CutMix&86.12$\pm$0.16\\
			\midrule
			ResNet-50+ST (ours)&85.72$\pm$0.11\\
			ResNet-50+MST (ours)&86.56$\pm$0.16\\
			ResNet-50+MDT (ours)&{\bfseries 86.72$\pm$0.12}\\
			\midrule
			\bottomrule[1.2pt]
		\end{tabular}
	\end{center}
	\caption{Performance of data augmentation methods on CUB-200-2011. MDT denotes Mixed Double Thumbnial. Both MDT and MST outperform CutMix. }
	\label{tab5}
	
\end{table}

\begin{figure}[h]
	\centering
	\includegraphics[width=0.98\columnwidth]{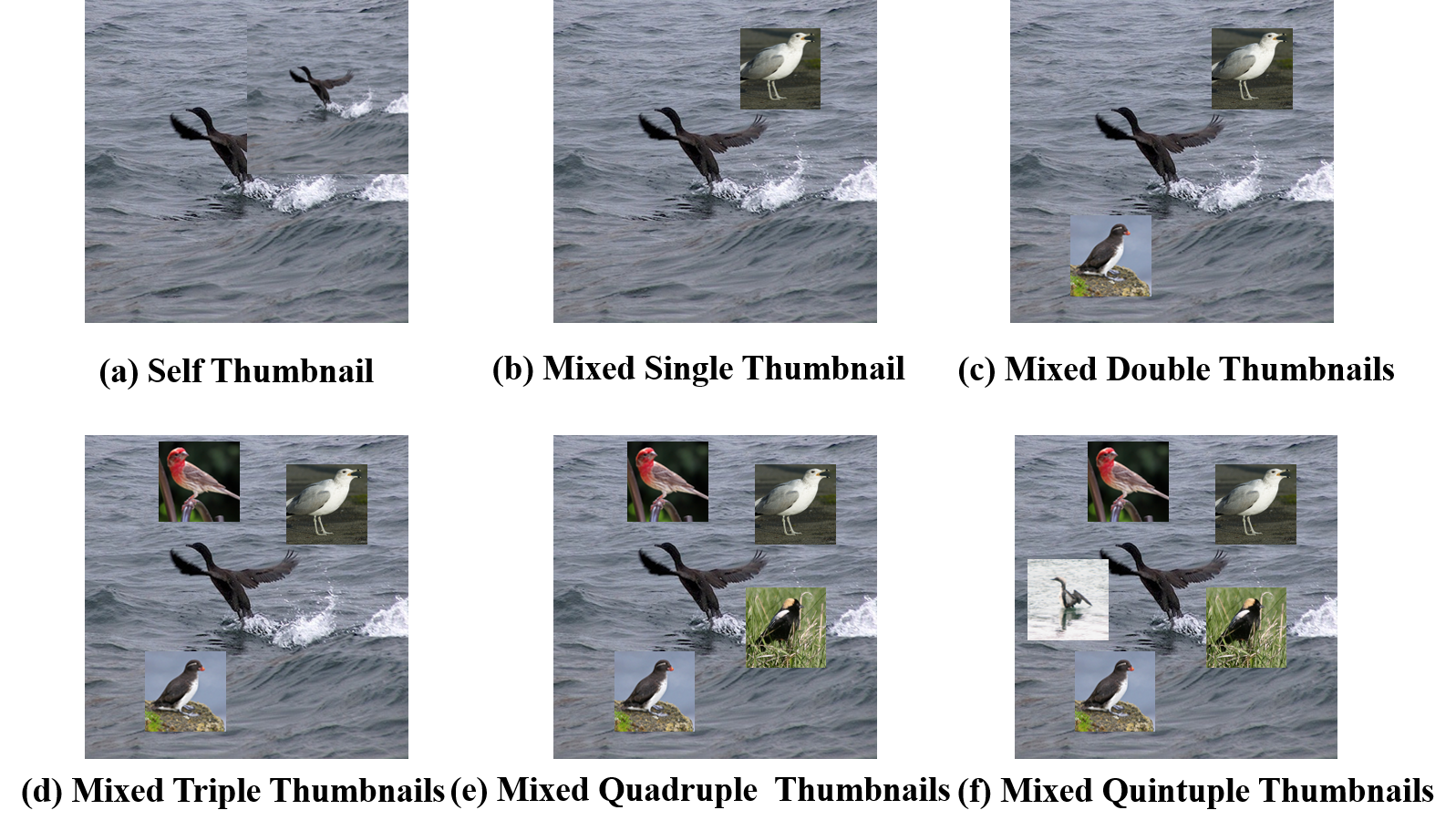} 
	\caption{Examples of training images using Mixed Multiple Thumbnails on CUB-200-2011.}
	\label{fig6}
\end{figure}

\begin{figure}[h]
	\centering
	\includegraphics[scale=0.34]{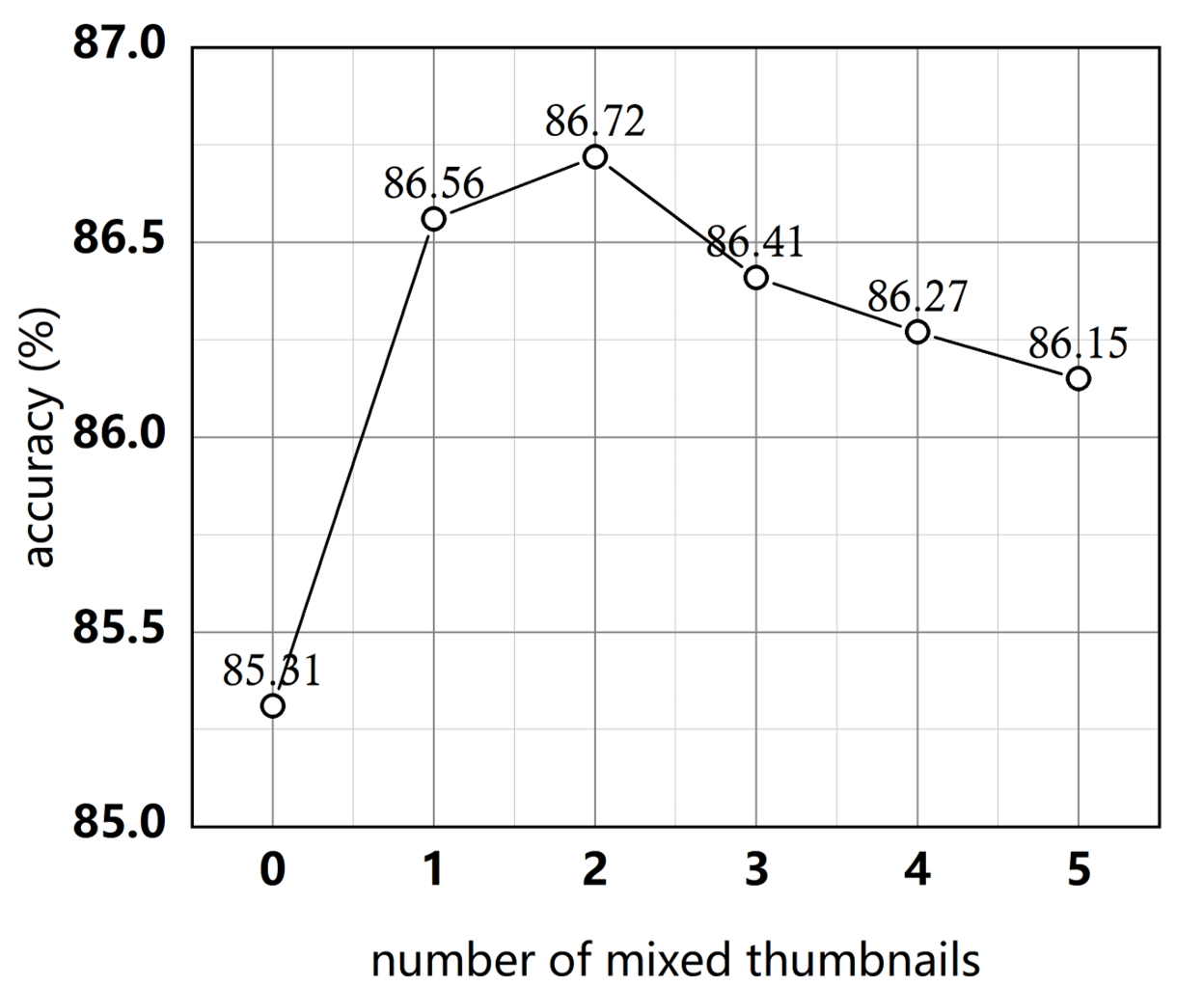}
	\caption{Performance of Mixed Multiple Thumbnails with different number of mixed thumbnails on CUB-200-2011. The network achieves the best performance when 2 thumbnails are pasted on each training image. }
	\label{fig7}
\end{figure}

\noindent \textbf{Performance of Mixed Multiple Thumbnails:} Unlike the above classification datasets, the number of each category's images in CUB-200-2011 is smaller, and each image's resolution is higher. This drives us to put more than one thumbnail on another image, as shown in Figure~\ref{fig6}. It is because putting more thumbnails on each image will not take up much semantics of the original image with such high resolution.  We set the thumbnail size to 130$\times$130, the weight of the origin image label $\lambda_1$ is set to 0.6, and the weight of the thumbnail label $\lambda_i$ is set to 0.2. As shown in Figure \ref{fig7}, with the increase in the number of thumbnails, the accuracy first rises and then falls and achieves the highest (86.62\%) when the number is two. Besides, we note that the original image still retains more than 60\% of area when  mixed with 5 thumbnails, so the network performance can still be significantly improved.

\noindent \textbf{Transfer Learning of Pretrained Model:} ImageNet pre-training is de-facto standard practice for many visual recognition tasks. We examine whether Cut-Thumbnail pre-trained models lead to better performances in fine-grained image classification. As shown in Table \ref{tab6}, the ResNet-50 with MST pre-trained model performs better than baseline. Furthermore, on the basis of taking the network pretrained by MST as the backbone, we have superimposed MST in the training, and the result achieves 87.76\%, which is more than 2.45\% improvement on the baseline.

\begin{table}[h]
	\begin{center}
		\begin{tabular}{p{5cm}p{1.8cm}}
			\toprule[1.2pt]
			\midrule
			Model&Accuracy(\%)\\
			\midrule
			ResNet-50 (baseline)&85.31$\pm$0.21\\
			ResNet-50+MST&86.56$\pm$0.16\\
			ResNet-50+MST pre-trained&86.17$\pm$0.12\\
			ResNet-50+MST pre-trained+MST&\bfseries{87.76$\pm$0.13}\\
			\midrule
			\bottomrule[1.2pt]
		\end{tabular}
	\end{center}
	\caption{Fine-grained image classification results on CUB-200-2011 with different backbone models.}
	\label{tab6}
\end{table}

\subsection{Object Detection in PASCAL VOC}
In this subsection, we show Cut-Thumbnail can also be applied for training object detector in Pascal VOC~\cite{everingham2010pascal} dataset. We use RetinaNet~\cite{lin2017focal} framework composed of a backbone network and two task-specific subnetworks for the experiments. The ResNet-50 backbone which is responsible for computing a convolutional feature map over an entire input image is initialized with ImageNet-pretrained model and then fine-tuned on Pascal VOC 2007 and 2012 trainval data. Models are evaluated on VOC 2007 test data using the mAP metric. We follow the fine-tuning strategy of the original method.

\begin{table}[h]
  \begin{tabular}{p{5cm}p{1.8cm}}
  \toprule[1.2pt]
  \midrule
        Model&mAP(\%)\\
  \midrule
        RetinaNet (baseline)&70.14$\pm$0.17\\
		RetinaNet+CutMix pre-trained&71.01$\pm$0.21\\
        RetinaNet+ST pre-trained (ours)&71.01$\pm$0.15\\
		RetinaNet+MST pre-trained (ours)&\bfseries{72.16$\pm$0.19}\\
  \midrule
  \bottomrule[1.2pt]
  \end{tabular}
  \caption{ Object detection results on Pascal VOC with RetinaNet.  The model pre-trained with MST achieves the best accuracy.}
  \label{tab7}
\end{table}

As shown in Table \ref{tab7}, the model pre-trained with MST achieves the best accuracy (72.16\%), +2.02\% higher than the baseline performance(70.14\%). It proves that our method is suitable for object detection task. Besides, CutMix works better than ST in Imagenet classification task, but the model pre-trained by ST performs equal to CutMix in object detection task. The results suggest that the model trained with Cut-Thumbnail can better capture the target objects.

\section{Related work}
\noindent \textbf{Regularization:} The regularization methods are effective for training neural networks. Dropout~\cite{srivastava2014dropout} injects noise into feature space by randomly zeroing the activation function to avoid overfitting. Besides, DropConnect~\cite{wan2013regularization}, Spatial Dropout~\cite{tompson2015efficient}, Droppath~\cite{larsson2016fractalnet}, DropBlock~\cite{ghiasi2018dropblock} and Weighted Channel Dropout~\cite{hou2019weighted} were also proposed as variants of Dropout. Besides, Batch Normalization~\cite{ioffe2015batch} improves the gradient propagation through network by normalizing the input for each layer.

\noindent \textbf{Data augmentation:} Data augmentation generates virtual training examples in the vicinity of the given training dataset to improve the generalization performance of network. Random cropping and horizontal flipping operatings~\cite{krizhevsky2012imagenet} are the most commonly used data augmentation techniques. By randomly removing contiguous sections of input images, Cutout~\cite{devries2017improved} improves the robustness of network. Random Erasing~\cite{zhong2020random} randomly selects a rectangle region in an image and erases its pixels with random values. Hide-and-Seek~\cite{provos2003hide} and GridMask~\cite{chen2020gridmask} can be regarded as upgraded versions of Cutout. AutoAugment~\cite{cubuk2018autoaugment} improves the inception-preprocess using reinforcement learning to search existing policies for the optimal combination.

\noindent \textbf{Mixed Sample Data Augmentation (MSDA):} Mixed Sample Data Augmentation has received increasing attention in recent years. Input mixup~\cite{zhang2017mixup} creates virtual training examples by linearly interpolating two input data and corresponding one-hot labels. Manifold mixup~\cite{verma2019manifold} is the variance of mixup, which encourages neural networks to predict less confidently on interpolations of hidden representations. Random image cropping and patching randomly~\cite{takahashi2018ricap} crops four images and patches them to create a new training image. Inspired by Cutout and Mixup, CutMix~\cite{yun2019cutmix} cut patches and pasted among training images. Based on CutMix, Attentive CutMix~\cite{walawalkar2020attentive}, FMix~\cite{harris2020fmix} and Puzzle Mix~\cite{kim2020puzzle} aim to capture the most important region(s) of one image and paste it(them) on another one. Cut-Thumbnail can be perfectly integrated with MSDA, because Cut-Thumbnail can introduce most semantics of one image to another image with occupying a small area of it.

\section{Conclusion}
We propose a simple, general and effective data augmentation method named Cut-Thumbnail, which is the first data augmentation method that introduces the idea of thumbnail to data augmentation strategy. We reduce an image to a small size and put it on itself or another image. Different strategies are designed to verify the effectiveness of the thumbnail, and finally Mixed Single Thumbnail works best on different visual tasks. On the ImageNet dataset, Cut-Thumbnail increases the baseline by 2.89\%. In fine-grained image classification, Cut-Thumbnail increases the accuracy of ResNet-50 from 85.31\% to 87.76\% on CUB-200-2011. In the task of Pascal VOC object detection, we improve the baseline by 2.02\% on RetinaNet. Extensive experiments have proved that Cut-Thumbnail makes the network better learn shape information and is suitable for different networks, datasets, and tasks. For future work, we plan to find better strategies and hyper-parameters for Cut-Thumbnail using reinforcement learning and apply Cut-Thumbnail to more types of visual tasks.

\bibliographystyle{ACM-Reference-Format}
\bibliography{re}


\begin{thebibliography}{42}


\ifx \showCODEN    \undefined \def \showCODEN     #1{\unskip}     \fi
\ifx \showDOI      \undefined \def \showDOI       #1{#1}\fi
\ifx \showISBNx    \undefined \def \showISBNx     #1{\unskip}     \fi
\ifx \showISBNxiii \undefined \def \showISBNxiii  #1{\unskip}     \fi
\ifx \showISSN     \undefined \def \showISSN      #1{\unskip}     \fi
\ifx \showLCCN     \undefined \def \showLCCN      #1{\unskip}     \fi
\ifx \shownote     \undefined \def \shownote      #1{#1}          \fi
\ifx \showarticletitle \undefined \def \showarticletitle #1{#1}   \fi
\ifx \showURL      \undefined \def \showURL       {\relax}        \fi
\providecommand\bibfield[2]{#2}
\providecommand\bibinfo[2]{#2}
\providecommand\natexlab[1]{#1}
\providecommand\showeprint[2][]{arXiv:#2}

\bibitem[\protect\citeauthoryear{Ballester and Araujo}{Ballester and
  Araujo}{2016}]%
        {ballester2016performance}
\bibfield{author}{\bibinfo{person}{Pedro Ballester} {and}
  \bibinfo{person}{Ricardo Araujo}.} \bibinfo{year}{2016}\natexlab{}.
\newblock \showarticletitle{On the performance of GoogLeNet and AlexNet applied
  to sketches}. In \bibinfo{booktitle}{\emph{Proceedings of the AAAI Conference
  on Artificial Intelligence}}, Vol.~\bibinfo{volume}{30}.
\newblock


\bibitem[\protect\citeauthoryear{Brendel and Bethge}{Brendel and
  Bethge}{2019}]%
        {brendel2019approximating}
\bibfield{author}{\bibinfo{person}{Wieland Brendel} {and}
  \bibinfo{person}{Matthias Bethge}.} \bibinfo{year}{2019}\natexlab{}.
\newblock \showarticletitle{Approximating cnns with bag-of-local-features
  models works surprisingly well on imagenet}.
\newblock \bibinfo{journal}{\emph{arXiv preprint arXiv:1904.00760}}
  (\bibinfo{year}{2019}).
\newblock


\bibitem[\protect\citeauthoryear{Chen, Papandreou, Kokkinos, Murphy, and
  Yuille}{Chen et~al\mbox{.}}{2017}]%
        {chen2017deeplab}
\bibfield{author}{\bibinfo{person}{Liang-Chieh Chen}, \bibinfo{person}{George
  Papandreou}, \bibinfo{person}{Iasonas Kokkinos}, \bibinfo{person}{Kevin
  Murphy}, {and} \bibinfo{person}{Alan~L Yuille}.}
  \bibinfo{year}{2017}\natexlab{}.
\newblock \showarticletitle{Deeplab: Semantic image segmentation with deep
  convolutional nets, atrous convolution, and fully connected crfs}.
\newblock \bibinfo{journal}{\emph{IEEE transactions on pattern analysis and
  machine intelligence}} \bibinfo{volume}{40}, \bibinfo{number}{4}
  (\bibinfo{year}{2017}), \bibinfo{pages}{834--848}.
\newblock


\bibitem[\protect\citeauthoryear{Chen}{Chen}{2020}]%
        {chen2020gridmask}
\bibfield{author}{\bibinfo{person}{Pengguang Chen}.}
  \bibinfo{year}{2020}\natexlab{}.
\newblock \showarticletitle{GridMask data augmentation}.
\newblock \bibinfo{journal}{\emph{arXiv preprint arXiv:2001.04086}}
  (\bibinfo{year}{2020}).
\newblock


\bibitem[\protect\citeauthoryear{Cubuk, Zoph, Mane, Vasudevan, and Le}{Cubuk
  et~al\mbox{.}}{2018}]%
        {cubuk2018autoaugment}
\bibfield{author}{\bibinfo{person}{Ekin~D Cubuk}, \bibinfo{person}{Barret
  Zoph}, \bibinfo{person}{Dandelion Mane}, \bibinfo{person}{Vijay Vasudevan},
  {and} \bibinfo{person}{Quoc~V Le}.} \bibinfo{year}{2018}\natexlab{}.
\newblock \showarticletitle{Autoaugment: Learning augmentation policies from
  data}.
\newblock \bibinfo{journal}{\emph{arXiv preprint arXiv:1805.09501}}
  (\bibinfo{year}{2018}).
\newblock


\bibitem[\protect\citeauthoryear{DeVries and Taylor}{DeVries and
  Taylor}{2017}]%
        {devries2017improved}
\bibfield{author}{\bibinfo{person}{Terrance DeVries} {and}
  \bibinfo{person}{Graham~W Taylor}.} \bibinfo{year}{2017}\natexlab{}.
\newblock \showarticletitle{Improved regularization of convolutional neural
  networks with cutout}.
\newblock \bibinfo{journal}{\emph{arXiv preprint arXiv:1708.04552}}
  (\bibinfo{year}{2017}).
\newblock


\bibitem[\protect\citeauthoryear{Diesendruck and Bloom}{Diesendruck and
  Bloom}{2003}]%
        {diesendruck2003specific}
\bibfield{author}{\bibinfo{person}{Gil Diesendruck} {and} \bibinfo{person}{Paul
  Bloom}.} \bibinfo{year}{2003}\natexlab{}.
\newblock \showarticletitle{How specific is the shape bias?}
\newblock \bibinfo{journal}{\emph{Child development}} \bibinfo{volume}{74},
  \bibinfo{number}{1} (\bibinfo{year}{2003}), \bibinfo{pages}{168--178}.
\newblock


\bibitem[\protect\citeauthoryear{Everingham, Van~Gool, Williams, Winn, and
  Zisserman}{Everingham et~al\mbox{.}}{2010}]%
        {everingham2010pascal}
\bibfield{author}{\bibinfo{person}{Mark Everingham}, \bibinfo{person}{Luc
  Van~Gool}, \bibinfo{person}{Christopher~KI Williams}, \bibinfo{person}{John
  Winn}, {and} \bibinfo{person}{Andrew Zisserman}.}
  \bibinfo{year}{2010}\natexlab{}.
\newblock \showarticletitle{The pascal visual object classes (voc) challenge}.
\newblock \bibinfo{journal}{\emph{International journal of computer vision}}
  \bibinfo{volume}{88}, \bibinfo{number}{2} (\bibinfo{year}{2010}),
  \bibinfo{pages}{303--338}.
\newblock


\bibitem[\protect\citeauthoryear{Gatys, Ecker, and Bethge}{Gatys
  et~al\mbox{.}}{2017}]%
        {gatys2017texture}
\bibfield{author}{\bibinfo{person}{Leon~A Gatys}, \bibinfo{person}{Alexander~S
  Ecker}, {and} \bibinfo{person}{Matthias Bethge}.}
  \bibinfo{year}{2017}\natexlab{}.
\newblock \showarticletitle{Texture and art with deep neural networks}.
\newblock \bibinfo{journal}{\emph{Current opinion in neurobiology}}
  \bibinfo{volume}{46} (\bibinfo{year}{2017}), \bibinfo{pages}{178--186}.
\newblock


\bibitem[\protect\citeauthoryear{Geirhos, Rubisch, Michaelis, Bethge, Wichmann,
  and Brendel}{Geirhos et~al\mbox{.}}{2018}]%
        {geirhos2018imagenet}
\bibfield{author}{\bibinfo{person}{Robert Geirhos}, \bibinfo{person}{Patricia
  Rubisch}, \bibinfo{person}{Claudio Michaelis}, \bibinfo{person}{Matthias
  Bethge}, \bibinfo{person}{Felix~A Wichmann}, {and} \bibinfo{person}{Wieland
  Brendel}.} \bibinfo{year}{2018}\natexlab{}.
\newblock \showarticletitle{ImageNet-trained CNNs are biased towards texture;
  increasing shape bias improves accuracy and robustness}.
\newblock \bibinfo{journal}{\emph{arXiv preprint arXiv:1811.12231}}
  (\bibinfo{year}{2018}).
\newblock


\bibitem[\protect\citeauthoryear{Ghiasi, Lin, and Le}{Ghiasi
  et~al\mbox{.}}{2018}]%
        {ghiasi2018dropblock}
\bibfield{author}{\bibinfo{person}{Golnaz Ghiasi}, \bibinfo{person}{Tsung-Yi
  Lin}, {and} \bibinfo{person}{Quoc~V Le}.} \bibinfo{year}{2018}\natexlab{}.
\newblock \showarticletitle{Dropblock: A regularization method for
  convolutional networks}. In \bibinfo{booktitle}{\emph{Advances in Neural
  Information Processing Systems}}. \bibinfo{pages}{10727--10737}.
\newblock


\bibitem[\protect\citeauthoryear{Girshick}{Girshick}{2015}]%
        {girshick2015fast}
\bibfield{author}{\bibinfo{person}{Ross Girshick}.}
  \bibinfo{year}{2015}\natexlab{}.
\newblock \showarticletitle{Fast r-cnn}. In
  \bibinfo{booktitle}{\emph{Proceedings of the IEEE international conference on
  computer vision}}. \bibinfo{pages}{1440--1448}.
\newblock


\bibitem[\protect\citeauthoryear{Harris, Marcu, Painter, Niranjan, and
  Hare}{Harris et~al\mbox{.}}{2020}]%
        {harris2020fmix}
\bibfield{author}{\bibinfo{person}{Ethan Harris}, \bibinfo{person}{Antonia
  Marcu}, \bibinfo{person}{Matthew Painter}, \bibinfo{person}{Mahesan
  Niranjan}, {and} \bibinfo{person}{Adam Pr{\"u}gel-Bennett~Jonathon Hare}.}
  \bibinfo{year}{2020}\natexlab{}.
\newblock \showarticletitle{FMix: Enhancing Mixed Sample Data Augmentation}.
\newblock \bibinfo{journal}{\emph{arXiv preprint arXiv:2002.12047}}
  (\bibinfo{year}{2020}).
\newblock


\bibitem[\protect\citeauthoryear{He, Gkioxari, Doll{\'a}r, and Girshick}{He
  et~al\mbox{.}}{2017}]%
        {he2017mask}
\bibfield{author}{\bibinfo{person}{Kaiming He}, \bibinfo{person}{Georgia
  Gkioxari}, \bibinfo{person}{Piotr Doll{\'a}r}, {and} \bibinfo{person}{Ross
  Girshick}.} \bibinfo{year}{2017}\natexlab{}.
\newblock \showarticletitle{Mask r-cnn}. In
  \bibinfo{booktitle}{\emph{Proceedings of the IEEE international conference on
  computer vision}}. \bibinfo{pages}{2961--2969}.
\newblock


\bibitem[\protect\citeauthoryear{He, Zhang, Ren, and Sun}{He
  et~al\mbox{.}}{2016}]%
        {he2016deep}
\bibfield{author}{\bibinfo{person}{Kaiming He}, \bibinfo{person}{Xiangyu
  Zhang}, \bibinfo{person}{Shaoqing Ren}, {and} \bibinfo{person}{Jian Sun}.}
  \bibinfo{year}{2016}\natexlab{}.
\newblock \showarticletitle{Deep residual learning for image recognition}. In
  \bibinfo{booktitle}{\emph{Proceedings of the IEEE conference on computer
  vision and pattern recognition}}. \bibinfo{pages}{770--778}.
\newblock


\bibitem[\protect\citeauthoryear{Hou and Wang}{Hou and Wang}{2019}]%
        {hou2019weighted}
\bibfield{author}{\bibinfo{person}{Saihui Hou} {and} \bibinfo{person}{Zilei
  Wang}.} \bibinfo{year}{2019}\natexlab{}.
\newblock \showarticletitle{Weighted channel dropout for regularization of deep
  convolutional neural network}. In \bibinfo{booktitle}{\emph{Proceedings of
  the AAAI Conference on Artificial Intelligence}}, Vol.~\bibinfo{volume}{33}.
  \bibinfo{pages}{8425--8432}.
\newblock


\bibitem[\protect\citeauthoryear{Ioffe and Szegedy}{Ioffe and Szegedy}{2015}]%
        {ioffe2015batch}
\bibfield{author}{\bibinfo{person}{Sergey Ioffe} {and}
  \bibinfo{person}{Christian Szegedy}.} \bibinfo{year}{2015}\natexlab{}.
\newblock \showarticletitle{Batch normalization: Accelerating deep network
  training by reducing internal covariate shift}.
\newblock \bibinfo{journal}{\emph{arXiv preprint arXiv:1502.03167}}
  (\bibinfo{year}{2015}).
\newblock


\bibitem[\protect\citeauthoryear{Kim, Choo, and Song}{Kim
  et~al\mbox{.}}{2020}]%
        {kim2020puzzle}
\bibfield{author}{\bibinfo{person}{Jang-Hyun Kim}, \bibinfo{person}{Wonho
  Choo}, {and} \bibinfo{person}{Hyun~Oh Song}.}
  \bibinfo{year}{2020}\natexlab{}.
\newblock \showarticletitle{Puzzle mix: Exploiting saliency and local
  statistics for optimal mixup}.
\newblock \bibinfo{journal}{\emph{arXiv preprint arXiv:2009.06962}}
  (\bibinfo{year}{2020}).
\newblock


\bibitem[\protect\citeauthoryear{Krizhevsky, Hinton, et~al\mbox{.}}{Krizhevsky
  et~al\mbox{.}}{2009}]%
        {krizhevsky2009learning}
\bibfield{author}{\bibinfo{person}{Alex Krizhevsky}, \bibinfo{person}{Geoffrey
  Hinton}, {et~al\mbox{.}}} \bibinfo{year}{2009}\natexlab{}.
\newblock \showarticletitle{Learning multiple layers of features from tiny
  images}.
\newblock  (\bibinfo{year}{2009}).
\newblock


\bibitem[\protect\citeauthoryear{Krizhevsky, Sutskever, and Hinton}{Krizhevsky
  et~al\mbox{.}}{2012}]%
        {krizhevsky2012imagenet}
\bibfield{author}{\bibinfo{person}{Alex Krizhevsky}, \bibinfo{person}{Ilya
  Sutskever}, {and} \bibinfo{person}{Geoffrey~E Hinton}.}
  \bibinfo{year}{2012}\natexlab{}.
\newblock \showarticletitle{Imagenet classification with deep convolutional
  neural networks}. In \bibinfo{booktitle}{\emph{Advances in neural information
  processing systems}}. \bibinfo{pages}{1097--1105}.
\newblock


\bibitem[\protect\citeauthoryear{Landau, Smith, and Jones}{Landau
  et~al\mbox{.}}{1988}]%
        {landau1988importance}
\bibfield{author}{\bibinfo{person}{Barbara Landau}, \bibinfo{person}{Linda~B
  Smith}, {and} \bibinfo{person}{Susan~S Jones}.}
  \bibinfo{year}{1988}\natexlab{}.
\newblock \showarticletitle{The importance of shape in early lexical learning}.
\newblock \bibinfo{journal}{\emph{Cognitive development}} \bibinfo{volume}{3},
  \bibinfo{number}{3} (\bibinfo{year}{1988}), \bibinfo{pages}{299--321}.
\newblock


\bibitem[\protect\citeauthoryear{Larsson, Maire, and Shakhnarovich}{Larsson
  et~al\mbox{.}}{2016}]%
        {larsson2016fractalnet}
\bibfield{author}{\bibinfo{person}{Gustav Larsson}, \bibinfo{person}{Michael
  Maire}, {and} \bibinfo{person}{Gregory Shakhnarovich}.}
  \bibinfo{year}{2016}\natexlab{}.
\newblock \showarticletitle{Fractalnet: Ultra-deep neural networks without
  residuals}.
\newblock \bibinfo{journal}{\emph{arXiv preprint arXiv:1605.07648}}
  (\bibinfo{year}{2016}).
\newblock


\bibitem[\protect\citeauthoryear{Lin, Goyal, Girshick, He, and Doll{\'a}r}{Lin
  et~al\mbox{.}}{2017}]%
        {lin2017focal}
\bibfield{author}{\bibinfo{person}{Tsung-Yi Lin}, \bibinfo{person}{Priya
  Goyal}, \bibinfo{person}{Ross Girshick}, \bibinfo{person}{Kaiming He}, {and}
  \bibinfo{person}{Piotr Doll{\'a}r}.} \bibinfo{year}{2017}\natexlab{}.
\newblock \showarticletitle{Focal loss for dense object detection}. In
  \bibinfo{booktitle}{\emph{Proceedings of the IEEE international conference on
  computer vision}}. \bibinfo{pages}{2980--2988}.
\newblock


\bibitem[\protect\citeauthoryear{Long, Shelhamer, and Darrell}{Long
  et~al\mbox{.}}{2015}]%
        {long2015fully}
\bibfield{author}{\bibinfo{person}{Jonathan Long}, \bibinfo{person}{Evan
  Shelhamer}, {and} \bibinfo{person}{Trevor Darrell}.}
  \bibinfo{year}{2015}\natexlab{}.
\newblock \showarticletitle{Fully convolutional networks for semantic
  segmentation}. In \bibinfo{booktitle}{\emph{Proceedings of the IEEE
  conference on computer vision and pattern recognition}}.
  \bibinfo{pages}{3431--3440}.
\newblock

\vfill\eject
\bibitem[\protect\citeauthoryear{Paszke, Gross, Chintala, Chanan, Yang, DeVito,
  Lin, Desmaison, Antiga, and Lerer}{Paszke et~al\mbox{.}}{2017}]%
        {paszke2017automatic}
\bibfield{author}{\bibinfo{person}{Adam Paszke}, \bibinfo{person}{Sam Gross},
  \bibinfo{person}{Soumith Chintala}, \bibinfo{person}{Gregory Chanan},
  \bibinfo{person}{Edward Yang}, \bibinfo{person}{Zachary DeVito},
  \bibinfo{person}{Zeming Lin}, \bibinfo{person}{Alban Desmaison},
  \bibinfo{person}{Luca Antiga}, {and} \bibinfo{person}{Adam Lerer}.}
  \bibinfo{year}{2017}\natexlab{}.
\newblock \showarticletitle{Automatic differentiation in pytorch}.
\newblock  (\bibinfo{year}{2017}).
\newblock


\bibitem[\protect\citeauthoryear{Provos and Honeyman}{Provos and
  Honeyman}{2003}]%
        {provos2003hide}
\bibfield{author}{\bibinfo{person}{Niels Provos} {and} \bibinfo{person}{Peter
  Honeyman}.} \bibinfo{year}{2003}\natexlab{}.
\newblock \showarticletitle{Hide and seek: An introduction to steganography}.
\newblock \bibinfo{journal}{\emph{IEEE security \& privacy}}
  \bibinfo{volume}{1}, \bibinfo{number}{3} (\bibinfo{year}{2003}),
  \bibinfo{pages}{32--44}.
\newblock


\bibitem[\protect\citeauthoryear{Ren, He, Girshick, and Sun}{Ren
  et~al\mbox{.}}{2015}]%
        {ren2015faster}
\bibfield{author}{\bibinfo{person}{Shaoqing Ren}, \bibinfo{person}{Kaiming He},
  \bibinfo{person}{Ross Girshick}, {and} \bibinfo{person}{Jian Sun}.}
  \bibinfo{year}{2015}\natexlab{}.
\newblock \showarticletitle{Faster r-cnn: Towards real-time object detection
  with region proposal networks}. In \bibinfo{booktitle}{\emph{Advances in
  neural information processing systems}}. \bibinfo{pages}{91--99}.
\newblock


\bibitem[\protect\citeauthoryear{Russakovsky, Deng, Su, Krause, Satheesh, Ma,
  Huang, Karpathy, Khosla, Bernstein, et~al\mbox{.}}{Russakovsky
  et~al\mbox{.}}{2015}]%
        {russakovsky2015imagenet}
\bibfield{author}{\bibinfo{person}{Olga Russakovsky}, \bibinfo{person}{Jia
  Deng}, \bibinfo{person}{Hao Su}, \bibinfo{person}{Jonathan Krause},
  \bibinfo{person}{Sanjeev Satheesh}, \bibinfo{person}{Sean Ma},
  \bibinfo{person}{Zhiheng Huang}, \bibinfo{person}{Andrej Karpathy},
  \bibinfo{person}{Aditya Khosla}, \bibinfo{person}{Michael Bernstein},
  {et~al\mbox{.}}} \bibinfo{year}{2015}\natexlab{}.
\newblock \showarticletitle{Imagenet large scale visual recognition challenge}.
\newblock \bibinfo{journal}{\emph{International journal of computer vision}}
  \bibinfo{volume}{115}, \bibinfo{number}{3} (\bibinfo{year}{2015}),
  \bibinfo{pages}{211--252}.
\newblock


\bibitem[\protect\citeauthoryear{Shi, Zhang, Dai, Zhu, Mu, and Wang}{Shi
  et~al\mbox{.}}{2020}]%
        {shi2020informative}
\bibfield{author}{\bibinfo{person}{Baifeng Shi}, \bibinfo{person}{Dinghuai
  Zhang}, \bibinfo{person}{Qi Dai}, \bibinfo{person}{Zhanxing Zhu},
  \bibinfo{person}{Yadong Mu}, {and} \bibinfo{person}{Jingdong Wang}.}
  \bibinfo{year}{2020}\natexlab{}.
\newblock \showarticletitle{Informative dropout for robust representation
  learning: A shape-bias perspective}.
\newblock \bibinfo{journal}{\emph{arXiv preprint arXiv:2008.04254}}
  (\bibinfo{year}{2020}).
\newblock


\bibitem[\protect\citeauthoryear{Srivastava, Hinton, Krizhevsky, Sutskever, and
  Salakhutdinov}{Srivastava et~al\mbox{.}}{2014}]%
        {srivastava2014dropout}
\bibfield{author}{\bibinfo{person}{Nitish Srivastava},
  \bibinfo{person}{Geoffrey Hinton}, \bibinfo{person}{Alex Krizhevsky},
  \bibinfo{person}{Ilya Sutskever}, {and} \bibinfo{person}{Ruslan
  Salakhutdinov}.} \bibinfo{year}{2014}\natexlab{}.
\newblock \showarticletitle{Dropout: a simple way to prevent neural networks
  from overfitting}.
\newblock \bibinfo{journal}{\emph{The journal of machine learning research}}
  \bibinfo{volume}{15}, \bibinfo{number}{1} (\bibinfo{year}{2014}),
  \bibinfo{pages}{1929--1958}.
\newblock


\bibitem[\protect\citeauthoryear{Szegedy, Liu, Jia, Sermanet, Reed, Anguelov,
  Erhan, Vanhoucke, and Rabinovich}{Szegedy et~al\mbox{.}}{2015}]%
        {szegedy2015going}
\bibfield{author}{\bibinfo{person}{Christian Szegedy}, \bibinfo{person}{Wei
  Liu}, \bibinfo{person}{Yangqing Jia}, \bibinfo{person}{Pierre Sermanet},
  \bibinfo{person}{Scott Reed}, \bibinfo{person}{Dragomir Anguelov},
  \bibinfo{person}{Dumitru Erhan}, \bibinfo{person}{Vincent Vanhoucke}, {and}
  \bibinfo{person}{Andrew Rabinovich}.} \bibinfo{year}{2015}\natexlab{}.
\newblock \showarticletitle{Going deeper with convolutions}. In
  \bibinfo{booktitle}{\emph{Proceedings of the IEEE conference on computer
  vision and pattern recognition}}. \bibinfo{pages}{1--9}.
\newblock


\bibitem[\protect\citeauthoryear{Takahashi, Matsubara, and Uehara}{Takahashi
  et~al\mbox{.}}{2018}]%
        {takahashi2018ricap}
\bibfield{author}{\bibinfo{person}{Ryo Takahashi}, \bibinfo{person}{Takashi
  Matsubara}, {and} \bibinfo{person}{Kuniaki Uehara}.}
  \bibinfo{year}{2018}\natexlab{}.
\newblock \showarticletitle{Ricap: Random image cropping and patching data
  augmentation for deep cnns}. In \bibinfo{booktitle}{\emph{Asian Conference on
  Machine Learning}}. \bibinfo{pages}{786--798}.
\newblock


\bibitem[\protect\citeauthoryear{Tompson, Goroshin, Jain, LeCun, and
  Bregler}{Tompson et~al\mbox{.}}{2015}]%
        {tompson2015efficient}
\bibfield{author}{\bibinfo{person}{Jonathan Tompson}, \bibinfo{person}{Ross
  Goroshin}, \bibinfo{person}{Arjun Jain}, \bibinfo{person}{Yann LeCun}, {and}
  \bibinfo{person}{Christoph Bregler}.} \bibinfo{year}{2015}\natexlab{}.
\newblock \showarticletitle{Efficient object localization using convolutional
  networks}. In \bibinfo{booktitle}{\emph{Proceedings of the IEEE conference on
  computer vision and pattern recognition}}. \bibinfo{pages}{648--656}.
\newblock


\bibitem[\protect\citeauthoryear{Verma, Lamb, Beckham, Najafi, Mitliagkas,
  Lopez-Paz, and Bengio}{Verma et~al\mbox{.}}{2019}]%
        {verma2019manifold}
\bibfield{author}{\bibinfo{person}{Vikas Verma}, \bibinfo{person}{Alex Lamb},
  \bibinfo{person}{Christopher Beckham}, \bibinfo{person}{Amir Najafi},
  \bibinfo{person}{Ioannis Mitliagkas}, \bibinfo{person}{David Lopez-Paz},
  {and} \bibinfo{person}{Yoshua Bengio}.} \bibinfo{year}{2019}\natexlab{}.
\newblock \showarticletitle{Manifold mixup: Better representations by
  interpolating hidden states}. In \bibinfo{booktitle}{\emph{International
  Conference on Machine Learning}}. PMLR, \bibinfo{pages}{6438--6447}.
\newblock


\bibitem[\protect\citeauthoryear{Wah, Branson, Welinder, Perona, and
  Belongie}{Wah et~al\mbox{.}}{2011}]%
        {wah2011caltech}
\bibfield{author}{\bibinfo{person}{Catherine Wah}, \bibinfo{person}{Steve
  Branson}, \bibinfo{person}{Peter Welinder}, \bibinfo{person}{Pietro Perona},
  {and} \bibinfo{person}{Serge Belongie}.} \bibinfo{year}{2011}\natexlab{}.
\newblock \showarticletitle{The caltech-ucsd birds-200-2011 dataset}.
\newblock  (\bibinfo{year}{2011}).
\newblock


\bibitem[\protect\citeauthoryear{Walawalkar, Shen, Liu, and
  Savvides}{Walawalkar et~al\mbox{.}}{2020}]%
        {walawalkar2020attentive}
\bibfield{author}{\bibinfo{person}{Devesh Walawalkar},
  \bibinfo{person}{Zhiqiang Shen}, \bibinfo{person}{Zechun Liu}, {and}
  \bibinfo{person}{Marios Savvides}.} \bibinfo{year}{2020}\natexlab{}.
\newblock \showarticletitle{Attentive Cutmix: An Enhanced Data Augmentation
  Approach for Deep Learning Based Image Classification}. In
  \bibinfo{booktitle}{\emph{ICASSP 2020-2020 IEEE International Conference on
  Acoustics, Speech and Signal Processing (ICASSP)}}. IEEE,
  \bibinfo{pages}{3642--3646}.
\newblock


\bibitem[\protect\citeauthoryear{Wan, Zeiler, Zhang, Le~Cun, and Fergus}{Wan
  et~al\mbox{.}}{2013}]%
        {wan2013regularization}
\bibfield{author}{\bibinfo{person}{Li Wan}, \bibinfo{person}{Matthew Zeiler},
  \bibinfo{person}{Sixin Zhang}, \bibinfo{person}{Yann Le~Cun}, {and}
  \bibinfo{person}{Rob Fergus}.} \bibinfo{year}{2013}\natexlab{}.
\newblock \showarticletitle{Regularization of neural networks using
  dropconnect}. In \bibinfo{booktitle}{\emph{International conference on
  machine learning}}. \bibinfo{pages}{1058--1066}.
\newblock


\bibitem[\protect\citeauthoryear{Yun, Han, Oh, Chun, Choe, and Yoo}{Yun
  et~al\mbox{.}}{2019}]%
        {yun2019cutmix}
\bibfield{author}{\bibinfo{person}{Sangdoo Yun}, \bibinfo{person}{Dongyoon
  Han}, \bibinfo{person}{Seong~Joon Oh}, \bibinfo{person}{Sanghyuk Chun},
  \bibinfo{person}{Junsuk Choe}, {and} \bibinfo{person}{Youngjoon Yoo}.}
  \bibinfo{year}{2019}\natexlab{}.
\newblock \showarticletitle{Cutmix: Regularization strategy to train strong
  classifiers with localizable features}. In
  \bibinfo{booktitle}{\emph{Proceedings of the IEEE International Conference on
  Computer Vision}}. \bibinfo{pages}{6023--6032}.
\newblock


\bibitem[\protect\citeauthoryear{Zagoruyko and Komodakis}{Zagoruyko and
  Komodakis}{2016}]%
        {zagoruyko2016wide}
\bibfield{author}{\bibinfo{person}{Sergey Zagoruyko} {and}
  \bibinfo{person}{Nikos Komodakis}.} \bibinfo{year}{2016}\natexlab{}.
\newblock \showarticletitle{Wide residual networks}.
\newblock \bibinfo{journal}{\emph{arXiv preprint arXiv:1605.07146}}
  (\bibinfo{year}{2016}).
\newblock


\bibitem[\protect\citeauthoryear{Zhang, Cisse, Dauphin, and Lopez-Paz}{Zhang
  et~al\mbox{.}}{2017}]%
        {zhang2017mixup}
\bibfield{author}{\bibinfo{person}{Hongyi Zhang}, \bibinfo{person}{Moustapha
  Cisse}, \bibinfo{person}{Yann~N Dauphin}, {and} \bibinfo{person}{David
  Lopez-Paz}.} \bibinfo{year}{2017}\natexlab{}.
\newblock \showarticletitle{mixup: Beyond empirical risk minimization}.
\newblock \bibinfo{journal}{\emph{arXiv preprint arXiv:1710.09412}}
  (\bibinfo{year}{2017}).
\newblock


\bibitem[\protect\citeauthoryear{Zhong, Zheng, Kang, Li, and Yang}{Zhong
  et~al\mbox{.}}{2020}]%
        {zhong2020random}
\bibfield{author}{\bibinfo{person}{Zhun Zhong}, \bibinfo{person}{Liang Zheng},
  \bibinfo{person}{Guoliang Kang}, \bibinfo{person}{Shaozi Li}, {and}
  \bibinfo{person}{Yi Yang}.} \bibinfo{year}{2020}\natexlab{}.
\newblock \showarticletitle{Random Erasing Data Augmentation.}. In
  \bibinfo{booktitle}{\emph{AAAI}}. \bibinfo{pages}{13001--13008}.
\newblock


\bibitem[\protect\citeauthoryear{Zhou, Khosla, Lapedriza, Oliva, and
  Torralba}{Zhou et~al\mbox{.}}{2016}]%
        {zhou2016learning}
\bibfield{author}{\bibinfo{person}{Bolei Zhou}, \bibinfo{person}{Aditya
  Khosla}, \bibinfo{person}{Agata Lapedriza}, \bibinfo{person}{Aude Oliva},
  {and} \bibinfo{person}{Antonio Torralba}.} \bibinfo{year}{2016}\natexlab{}.
\newblock \showarticletitle{Learning deep features for discriminative
  localization}. In \bibinfo{booktitle}{\emph{Proceedings of the IEEE
  conference on computer vision and pattern recognition}}.
  \bibinfo{pages}{2921--2929}.
\newblock


\end{thebibliography}

\end{document}